\documentclass[11pt,a4paper]{article}
\usepackage[hyperref]{naaclhlt2019}
\usepackage{times}
\usepackage{latexsym}
\usepackage{amsmath, amssymb}
\usepackage{txfonts}
\usepackage{booktabs}
\usepackage{comment}
\usepackage{color}
\usepackage{url}
\usepackage{graphicx}  
\usepackage{color}

\aclfinalcopy 
\def\aclpaperid{961} 

\title{Graph Pattern Entity Ranking Model \\for Knowledge Graph Completion}

\author{Takuma Ebisu${}^\text{1,2}$ \\
  Affiliation / Address line 1 \\
  Affiliation / Address line 2 \\
  Affiliation / Address line 3 \\
  {\tt email@domain} \\\And
  Second Author \\
  Affiliation / Address line 1 \\
  Affiliation / Address line 2 \\
  Affiliation / Address line 3 \\
  {\tt email@domain} \\}
\author{Takuma Ebisu${}^\text{1,2}$\and Ryutaro Ichise${}^\text{2,1,3}$\\
	${}^\text{1}$SOKENDAI, Chiyoda-ku, Tokyo, Japan\\
	${}^\text{2}$National Institute of Informatics, Chiyoda-ku, Tokyo, Japan\\
	${}^\text{3}$National Institute of Advamced Industrial Science and Technology, Koto-ku, Tokyo, Japan\\
	\{takuma,ichise\}@nii.ac.jp\\
}

\date{}

\begin{document}
\maketitle
\begin{abstract}
Knowledge graphs have evolved rapidly in recent years and their usefulness has been demonstrated in many artificial intelligence tasks.
However, knowledge graphs often have lots of missing facts.
To solve this problem, many knowledge graph embedding models have been developed to populate knowledge graphs and these have shown outstanding performance.
However, knowledge graph embedding models are so-called black boxes, and the user does not know how the information in a knowledge graph is processed and the models can be difficult to interpret.
In this paper, we utilize graph patterns in a knowledge graph to overcome such problems.
Our proposed model, the {\it graph pattern entity ranking model} (GRank), constructs an entity ranking system for each graph pattern and evaluates them using a ranking measure.
By doing so, we can find graph patterns which are useful for predicting facts.
Then, we perform link prediction tasks on standard datasets to evaluate our GRank method.
We show that our approach outperforms other state-of-the-art approaches such as ComplEx and TorusE for standard metrics such as HITS@{\it n} and MRR.
Moreover, our model is easily interpretable because the output facts are described by graph patterns. 
\end{abstract}
	\section{Introduction}
Knowledge graphs can be used to describe real-world relations as facts in a form that a computer can easily process 
{and has been used for many artificial intelligence tasks \cite{Hakimov:2012:NER:2237867.2237871,Daiber:2013:IEA:2506182.2506198,D14-1067}.}
In a knowledge graph, a fact is represented by a labeled and directed edge, called a triple $(h,r,t)$, where $h$ and $t$ are entity nodes and $r$ is a relation label of an edge from $h$ to $t$.
Knowledge graphs such as YAGO \cite{DBLP:conf/www/SuchanekKW07}, DBpedia \cite{DBLP:conf/semweb/AuerBKLCI07}, and Freebase \cite{Bollacker:2008:FCC:1376616.1376746} have developed rapidly in recent years and are used for many artificial intelligence tasks such as question answering, content tagging, fact-checking, and knowledge inference.
Although some knowledge graphs already contain millions of entities and billions of facts, they might still be incomplete and some facts may be missing.
Hence, we need to develop a system that can predict missing facts to complete knowledge graphs automatically.

Many kinds of models for {\it link prediction} have been developed to estimate unknown facts.
Knowledge graph embedding models, which are the most widely used approach in this field, map entities and relations in a knowledge graph onto a vector space and obtain the latent underlying features.
However, these models are generally difficult to interpret, as we do not know how information is processed in the models and the predicted facts are output without explanation.

In this paper, we construct statistical models based on {\it graph pattern matching}.
These models are not only easy to interpret compared to knowledge graph embedding models but also outperform state-of-the-art models for link prediction.

Our main contributions in this paper are as follows:
\begin{itemize}
	\item Defining {\it graph pattern association rules} (GPARs) for a knowledge graph.
	\item Introducing {\it a graph pattern probability model} (GPro) and discussing its flaws.
	\item Proposing a novel model, {\it the graph pattern entity ranking model} (GRank), which uses graph patterns to rank entities.
	\item Proposing distributed rankings to address the problem arising from having the same score for multiple entities.
	\item Evaluating the proposed models through link prediction tasks for standard datasets:
	It is shown that our model outperforms most state-of-the-art knowledge graph embedding models for the HITS@{\it n} and MRR metrics.
\end{itemize}

The remainder of this paper is organized as follows.
In Section \ref{relatedwork}, we discuss related work on link prediction.
In Section \ref{preliminaries}, we define the terms and notation used in this paper.
In Section \ref{standard}, we define standard confidences for GPARs and discuss their problems.
In Section \ref{proposing}, we propose the GRank model to deal with these problems.
In Section \ref{experiments}, we present an experimental study in which we compare our models with baseline results for benchmark datasets.
In Section \ref{conclusions}, we conclude this paper.
\section{Related Work}
\label{relatedwork}
We categorize related work for link prediction into two groups: work on knowledge graph embedding models (which are latent feature models) and work on observed feature models.

\subsection{Knowledge Graph Embedding Models}
Recently, knowledge graph embedding models have yielded great results in link prediction tasks.
Knowledge graph embeddings models embed entities and relations on a continuous space and can be roughly classified into three types: translation-based models, bilinear models, and neural network-based models. 

The first translation-based model was the TransE \cite{DBLP:conf/nips/BordesUGWY13} model, which gained attention because of its effectiveness and simplicity.
TransE employs the principle $\boldsymbol{h}+\boldsymbol{r}= \boldsymbol{t}$, where $\boldsymbol{h},\boldsymbol{r}$ and $\boldsymbol{t}$ are the embeddings of $h,r$ and $t$, respectively.
While this principle efficiently captures first-order rules, the TransE approach still has some problems.
The conflict between principle and regularization is one of these problems and the TorusE \cite{DBLP:journals/corr/abs-1711-05435} model was recently proposed to solve this problem by embedding entities and relations on a torus manifold.

RESCAL \cite{DBLP:conf/icml/NickelTK11} was the first bilinear model, where
each relation is represented by a square matrix and the score of the triple $(h,r,t)$ is calculated by a bilinear map which corresponds to the matrix of the relation $r$ and whose arguments are $\boldsymbol{h}$ and $\boldsymbol{t}$.
Hence, RESCAL represents the most general form of a bilinear model.
Extensions of RESCAL have been proposed by restricting bilinear functions, for example,
DistMult \cite{DBLP:journals/corr/YangYHGD14a} and ComplEx \cite{DBLP:conf/icml/TrouillonWRGB16} restrict the matrices representing the relations to diagonal matrices.

Neural network-based models have layers and an activation function like a neural network.
The Neural Tensor Network (NTN) \cite{DBLP:conf/nips/SocherCMN13} has a standard linear neural network structure and a bilinear tensor structure, and
can be considered as a generalization of RESCAL, where the weight of the network is trained for each relation.
Graph Convolutional Networks (GCNs) \cite{NIPS2015_5954,NIPS2016_6081,DBLP:journals/corr/KipfW16} exploit the convolution operator to capture local information for a graph, however these models are for undirected graphs.
Relational GCNs \cite{DBLP:journals/corr/SchlichtkrullKB17} and ConvE \cite{DBLP:journals/corr/DettmersMSR17} are generalizations of GCNs for knowledge graphs. 

Knowledge graph embedding is the standard approach for link prediction.
However, it suffers from low interpretability, resulting in triples which are predicted without any clear reason.

\subsection{Observed Feature Models}
The main advantage of observed feature models over knowledge graph embedding models is their interpretability. 
Additionally, Toutanova et al. \shortcite{observed-versus-latent-features-for-knowledge-base-and-text-inference} proposed a relatively simple logistic regression model, the Node+LinkFeat model, which utilizes only one-hop information in a knowledge graph and demonstrated that it performs far better for link prediction on standard datasets than most existing knowledge graph embedding models. 
However, it has also been shown that the Node+LinkFeat model cannot deal with a low-redundancy dataset because the model uses information which is too local.
On the other hand, it has shown that a logistic regression model, the PRA model \cite{DBLP:journals/ml/LaoC10,DBLP:conf/emnlp/LaoMC11}, which utilizes multi-hop information do not have sufficient accuracy \cite{Liu:2016:HRW:2911451.2911509}.
This suggests logistic regression does not have enough power to deal with deep information.
These studies have motivated research toward developing a more efficient model utilizing deeper information.
\begin{figure*}[tb]
	\centering
	\includegraphics[scale=1,trim=20 0 0 0]{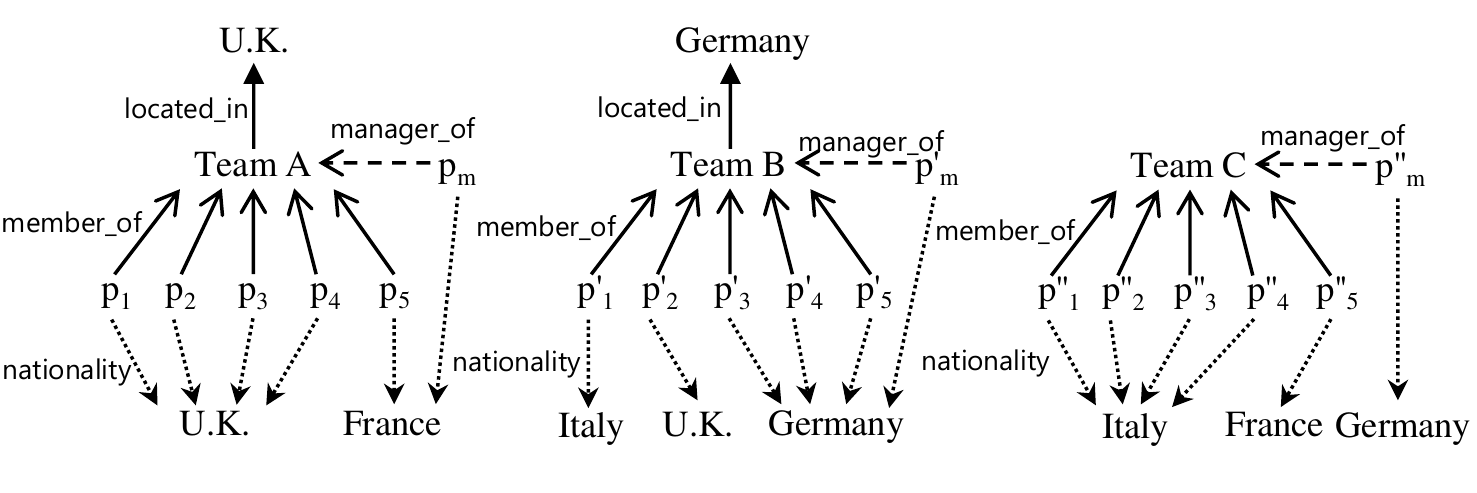}
	\caption{Graph $G_{ex}$ of sports teams.
	}
	\label{graph}
\end{figure*}
\begin{figure*}[tb]
	\centering
	\includegraphics[scale=1,trim=10 0 0 0]{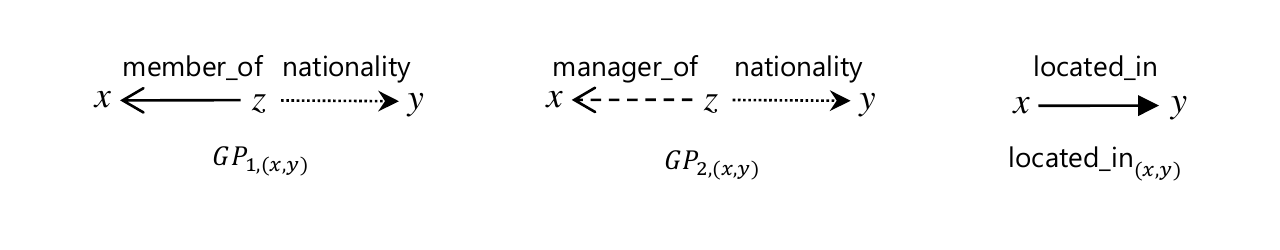}
	\caption{Graph patterns on $G_{ex}$.
	}
	\label{pattern}
\end{figure*}

We begin by discussing GPARs, which were proposed recently by \citeauthor{DBLP:journals/pvldb/FanWWX15} \shortcite{DBLP:journals/pvldb/FanWWX15}, and have shown their usefulness for social network graphs because graph patterns can capture deeper information lying in a knowledge graph and a GPAR explicitly describe the process of prediction.
However, the definition of GPARs by \citeauthor{DBLP:journals/pvldb/FanWWX15} cannot be applied to a knowledge graph because \citeauthor{DBLP:journals/pvldb/FanWWX15} assumes a different structure for a social network graph than a knowledge graph.
In the following section, we define GPARs for a knowledge graph.

\section{Preliminaries}
\label{preliminaries}
In this section, we introduce the definitions and notation required to discuss GPAR-based models.
\subsection{Graph Pattern Association Rules on Knowledge Graphs}
\label{definitions}

We modify GPARs for application to a knowledge graph following the definitions of \citeauthor{DBLP:journals/pvldb/FanWWX15} \shortcite{DBLP:journals/pvldb/FanWWX15}.

{\bf Knowledge Graph:} A {\it graph} is defined as $G=\{(h,r,t)\}\subset E\times R \times E$, where $E$ denotes a set of entities and $R$ denotes a set of relations. An element $(h,r,t)$ of $G$ is called a triple and represents the directed relation $r$ between $h$ and $t$.

An example graph $G_{ex}$
is shown in Figure \ref{graph}, where ${p_i}$ represents a person, Teams ``A", ``B", and ``C" represent sports teams, and countries are entities in $E_{ex}$ with labeled arrows between two entities representing directed relations in $R_{ex}$.

{\bf Graph Pattern:} A {\it graph pattern} on $G$ is a graph $GP_{(x,y)}=\{(z_i,r,z_j)\}\subset  V_{GP}\times R \times V_{GP}$, where $V_{GP}$ denotes a set of variables, $x$ and $y$ are two designated variables, and $R$ is the set of relations of $G$. We suppose $V_GP$ has no redundancy, in other words, $\forall z \in V_{GP}, \exists (z_i,r,z_j) \in GP_{(x,y)}, z=z_i \vee z=z_j$. 

Some examples of graph patterns on $G_{ex}$ are shown in Figure \ref{pattern}, where $GP_{1,(x,y)}=\{(z,{\sf member\_of},x),(z,{\sf nationality},y)\}$, $GP_{2,(x,y)}=\{(z,{\sf manager\_of},x),(z,{\sf nationality},y)\}$, and ${\sf located\_in}_{(x,y)}=\{(x,{\sf located\_in},y)\}$.
{Our focus in this paper is on finding useful graph patterns for link prediction.}

{\bf Graph Pattern Matching Function:} A {\it matching function} of $GP_{(x,y)}$ on $(h,t) \in E \times E$ is an injective function $m:V_{GP} \rightarrow E$ that satisfies the following conditions: $m(x)=h$, $m(y)=t$, and for all $(z_i,r,z_j) \in GP_{(x,y)}$, $(m(z_i),r,m(z_j))\in G$. $M(GP_{(x,y)},(h,t))$ denotes the set of all matching functions of $GP_{(x,y)}$ on $(h,t)$. We say $GP_{(x,y)}$ matches $(h,t)$ if there is at least one matching function of $GP_{(x,y)}$ on $(h,t)$ (i.e. $M(GP_{(x,y)},(h,t))\neq \emptyset$).

For example, $m:V_{GP_{1,(x,y)}}\rightarrow \ {E_{ex}}$ ($m(x)={\rm Team\ A}, m(z)={\rm p_1}, m(y)={\rm U.K.}$) is a matching function of $GP_{1,(x,y)}$ on $({\rm Team\ A},{\rm U.K.})$.

{\bf GPAR:} A {\it graph pattern association rule} (GPAR) $AR$ is defined as $GP_{(x,y)} \Rightarrow r_{(x,y)}$, where $GP_{(x,y)}$ and $r_{(x,y)}$ are graph patterns and $r_{(x,y)} = \{(x,r,y)\}$.

For example, 	a GPAR $AR_1=P_{1,(x,y)}\Rightarrow {\sf located\_in}_{(x,y)}$ would indicate that if there is a matching function of $GP_{1,(x,y)}$ on $(h,t)$, then it is likely that there is also a matching function of ${\sf located\_in}_{(x,y)}$  on $(h,t)$, i.e. $(h,{\sf located\_in},t)$ is a fact.


\subsection{Reconstruction of Knowledge Graph to Queries}
\label{reconstruction}
Our task is the {\it link prediction} of a knowledge graph, i.e. to predict the missing entity of a {\it query}, which is formally defined as follows:

{\bf Query:} A {\it query} is a triple which is missing an entity: $(h,r,?)$ or $(?,r,t)$.

We divide a knowledge graph $G$ into queries and answers to use as training data for our model.
Let $Q_{r,{\rm head}}$ ($Q_{r,{\rm tail}}$) denote the set of training queries missing a head (tail) entity for a relation $r$ obtained from $G$; then $Q_{r,{\rm head}}$ ($Q_{r,{\rm tail}}$) is defined as follows:
\begin{eqnarray*}
Q_{r,{\rm head}}=\{(?,r,t)\mid(h,r,t)\in G\},\\ \ Q_{r,{\rm tail}}=\{(h,r,?)\mid(h,r,t)\in G\}
\end{eqnarray*}
In this case, the answers of training queries are defined as follows:
\begin{eqnarray*}
	a_{(?,r,t)}=\{h\mid (h,r,t)\in G\}, \\a_{(h,r,?)}=\{t\mid (h,r,t)\in G\}
\end{eqnarray*}
A knowledge graph usually contains only positive triples.
Hence, we adopt the {\it partial completeness assumption} (PCA) \cite{DBLP:conf/www/GalarragaTHS13,DBLP:journals/vldb/GalarragaTHS15} to generate negative answers.

{\bf Partial Completeness Assumption:} if $(h,r,t)$ is in $G$, then
\begin{eqnarray}
\forall t'\in E, ((h,r,t')\notin G \Rightarrow (h,r,t') {\rm \ is \ negative})\\
\forall h'\in E, ((h',r,t)\notin G \Rightarrow (h',r,t) {\rm \ is \ negative})
\end{eqnarray}
The standard PCA definition consists only of Equation (1), but we add Equation (2) because we also need to allow negative answers for $Q_{r,{\rm head}}$.
Under PCA, negative answers for each question are defined as follows:
\[n_{(?,r,t)}=E\setminus a_{(?,r,t)},\ n_{(h,r,?)}=E\setminus a_{(h,r,?)}
\]
\section{Standard Confidence and Problems}
\label{standard}
\subsection{AMIE with GPARs}
\label{AMIE}
An association rule is essentially a binary classifier, i.e. the antecedent of an association rule matches or does not match, and an association rule is thus evaluated.
Following this idea, we suggest the most straightforward way to define the {\it confidence}, which indicates the reliability of an association rule, is the conditional probability, which is the probability of the consequent given the antecedent for a GPAR.
The conditional probability $Pr_{tail}(r_{(x,y)}\mid GP_{(x,y)})$ of a GPAR $GP_{(x,y)} \Rightarrow r_{(x,y)}$ to predict a tail is defined as follows:
\begin{equation*}
\begin{aligned}
&{\it conf}_{tail}(GP_{(x,y)} \Rightarrow r_{(x,y)})=Pr_{tail}(r_{(x,y)}\mid GP_{(x,y)})=\\
&\frac{\sum_{(h,r,?)\in Q_{r,{\rm tail}}}|\{t \in a_{(h,r,?)}\mid M(GP_{(x,y)} ,(h,t))\neq \emptyset  \}|}
{\sum_{(h,r,?)\in Q_{r,{\rm tail}}}|\{t' \in E \mid M(GP_{(x,y)},(h,t'))\neq \emptyset \}|}
\end{aligned}
\end{equation*}
{For each query, the candidate entities found by the graph pattern are counted for the denominator while only correct entities are counted for the numerator.}
This confidence is used to evaluate GPARs only to answer queries with a missing tail because $Q_{r_{tail}}$ and its answers are used to define it.

Interestingly, GPARs with this confidence are equivalent to AMIE \cite{DBLP:conf/www/GalarragaTHS13,DBLP:journals/vldb/GalarragaTHS15}, which was proposed to find horn clauses for a knowledge graph, although 
AMIE was proposed before the appearance of GPARs.
However, AMIE originally has only one confidence value for a GPAR because AMIE is not designed for link prediction.
Hence, we introduce the following alternative definition for the confidence value to answer a query missing a head entity.
\subsection{Standard Link Prediction Model and Problems}
\label{problems}
We define another confidence to deal with a query with a missing head entity as follows:
\begin{equation*}
\begin{aligned}
&{\it conf}_{head}(GP_{(x,y)} \Rightarrow r_{(x,y)})=Pr_{head}(r_{(x,y)}\mid GP_{(x,y)})=\\
&\frac{\sum_{(?,r,t)\in Q_{r,{\rm head}}}|\{h \in a_{(?,r,t)}\mid M(GP_{(x,y)} ,(h,t))\neq \emptyset  \}|}
{\sum_{(?,r,t)\in Q_{r,{\rm head}}}|\{h'\in E\mid M(GP_{(x,y)},(h',t))\neq \emptyset  \}|}
\end{aligned}
\end{equation*}
Additionally, we restrict matching functions to injective functions as defined in Section \ref{definitions}, which is different from AMIE, because the restriction avoids redundant matching functions which map multiple variables to the same entity and gives a good bias for real-world knowledge.
For example, an GPAR $GP_{3,(x,y)} \Rightarrow {\sf sibling\_of}_{(x,y)}$, where $GP_{3,(x,y)}=\{(z,{\sf parent\_of},x),(y,{\sf child\_of},z)\}$, is helpful to predict siblings.
However, let $p$ represent a person, $GP_{3,(x,y)}$ matches $(p,p)$ although $p$ is not a sibling of $p$. 
The above restriction omits such concerns.
For another example, a GPAR $\{(z_1,{\sf manager\_of},x),\allowbreak (z_1,{\sf manager\_of},z_2),\allowbreak (z_2,{\sf located\_in},y)\}\allowbreak \Rightarrow {\sf located\_in}_{(x,y)}$ on the graph $G_{ex}$ in Figure \ref{graph} should not be considered helpful because $m(x)=m(z_2)$ holds for a matching function $m$ of the antecedent pattern and as a result, the GPAR is almost tautological.
We consider two confidence values for GPARs, $conf_{tail}$ and $conf_{head}$, referred to as the {\it graph pattern probability model} (GPro).

However, GPro cannot deal with queries where counting the number of matching functions is crucial.
An example where the number of matching functions is important is shown in Figure \ref{graph}.
In $G_{ex}$, the country that Team C is located in is missing.
One might guess that Team C is located in Italy because most of the Team C players have Italian nationality and the nationality of a player often matches the country that the team is located in.
However, GPro underestimates the GPAR $AR_{1,(x,y)}=GP_{1,(x,y)}\Rightarrow {\sf located\_in}_{(x,y)}$, which is equivalent to one's guessing process:
${\it conf}_{tail}(AR_{1,(x,y)})=2/5$, while ${\it conf}_{tail}(AR_{2,(x,y)})=1/2$, where $AR_{2,(x,y)}=GP_{2,(x,y)}\Rightarrow {\sf located\_in}_{(x,y)}$.
Hence, GPro judges $AR_{2,(x,y)}$ is more useful than $AR_{1,(x,y)}$, and
as a result, GPro predicts Team C is located in Germany rather than Italy.

This problem is caused by considering a GPAR as a binary classifier, i.e. the matching number is not taken into account.	
For example, if we apply $AR_{1,(x,y)}=P_{1,(x,y)}\Rightarrow {\sf located\_in}_{(x,y)}$ to a query $({\rm Team\ A},{\sf located\_in},?)$ in the traditional way (as a binary classifier), the output will contain two entities with equal weighting, the U.K. and France, because $P_{1,(x,y)}$ matches $({\rm Team\ A},{\rm U.K.})$ and $({\rm Team\ A},{\rm France})$.	
Then, one of the output entities is correct and the other is incorrect.	
This is the reason why $AR_{1,(x,y)}$ is underestimated.

To deal with this problem, in this paper, we consider a GPAR as an entity ranking system by counting the number of matching functions of the antecedent graph pattern rather than considering as a binary classifier.

\section{GPAR as Entity Ranking System and Evaluation Metrics}
\label{proposing}
As well as considering a GPAR as a binary classifier, we consider it as an entity ranking system.
Entities are ranked according to a score, based on their number of matching functions.

Moreover, we introduce the {\it distributed rankings} for entities, which are proposed to deal with situations where multiple entities have the same score.
Then, we define the evaluation metrics for the distributed rankings to evaluate GPARs for link prediction.

These approaches overcome the problems shown in Section \ref{problems}.

\subsection{GPAR as Entity Ranking System}
\label{rankingsystem}
We consider a GPAR as a ranking system in this section to rank queries for which counting the number of matching functions of the antecedent is helpful, as shown in Section \ref{problems}.

First, we define a scoring function whose arguments are a graph pattern $GP_{(x,y)}$ and a pair of entities $(h,t)$.
The scoring function returns the number of matching functions of a pattern on a pair, which is formally defined as follows:
\[
score(GP_{(x,y)},(h,t))=|M(GP_{(x,y)},(h,t))|
\]
Given a pattern $GP_{(x,y)}$ and a query $(h,r,?)$, we can obtain the $score(GP_{(x,y)},(h,t'))$ for each candidate tail entity $t'$.
Then we obtain the rankings of the tail entities in descending order of the scores.
The head entity rankings for a query $(?,r,t)$ are also obtained in this way.
This ranking method gives us new perspective when we apply GPARs to answer a query.
For example, if we apply $AR_{1,(x,y)}=P_{1,(x,y)}\Rightarrow {\sf located\_in}_{(x,y)}$ to a query $({\rm Team\ A},{\sf located\_in},?)$ the U.K. will be ranked first and France second.
In this situation, we can say that $AR_{1,(x,y)}$ works because the correct entity ranks higher than the wrong entity.  
We can basically evaluate a GPAR as an entity ranking system by evaluating output rankings by an evaluation metric for an ranking system such as the {\it mean average precision}.
However, often multiple entities have the same score and traditional metrics cannot deal with the situation.
To deal with this problem, we propose a new concept, called {\it distributed rankings}, and the corresponding metrics in the following sections.

\subsection{Distributed Rankings}
We propose distributed rankings where each entity can distribute over multiple ranks and each rank can have multiple entities, to deal with situations where multiple entities have the same score.

Traditional rankings of entities are represented by a matrix $Rank=(rank_{i,j})\in \{0,1\}^{n\times n}$, where $n$ is the number of entities, and for each column and row there is one $1$ element. 
In this matrix, columns represent entities and rows represent ranks. For example, $rank_{i,j}=1$ means that the entity $j$ has rank $i$.
On the other hand, distributed rankings of entities are represented by a matrix $dRank=(drank_{i,j})\in [0,1]^{n\times n}$, where the summation of a column or a row is equal to $1$.
Different from traditional rankings, the value of each element is continuous and multiple elements can be greater than 0 in a column or a row.
For example, $rank_{i,j}=0.5$ means that half of the entity $j$ has rank $i$.
Note that a traditional ranking matrix is a distributed ranking matrix.

Given a pattern $GP_{(x,y)}$ and a query $(h,r,?)$, We obtain distributed rankings of entities, $dRANK({GP_{(x,y)},(h,r,?)})$, according to their scores as follows.
Let $a$ be the number of entities whose scores are greater than the entity represented by $j$ and let $b$ be the number of entities whose scores are the same as the entity represented by $j$. Then, $drank_{i,j}$, an element of $dRANK({GP_{(x,y)},(h,r,?)})$, is determined to be $1/b$ for $a+1\leqq i \leqq a+b$ and $0$ otherwise.
Distributed rankings of head entities for a query $(?,r,t)$ are obtained in the same way, and we refer to them as $dRANK({GP_{(x,y)},(?,r,t)})$.
Unlike traditional rankings, distributed rankings are uniquely determined from the scores of entities.

Traditional rankings can be evaluated by metrics such as the average precision or the cumulative gain.
However, distributed rankings cannot be evaluated by these metrics.
Hence, we require a different evaluation metric for distributed rankings.


\subsection{Evaluation of GPARs as Entity Ranking System}
\label{realGRank}
We use a GPAR to obtain distributed entity rankings as shown in Section \ref{rankingsystem}.
In this section, we define a metric to evaluate distributed rankings of entities by generalizing the {\it average precision} to evaluate a GPAR.

For a pattern $GP_{(x,y)}$ and a training query $(h,r,?)$, the {\it distributed precision at} $k$, $dPre_k$, of $dRANK({GP_{(x,y)},(h,r,?)})$ is defined as follows:
\begin{equation*}
\begin{aligned}
dPre_k(&{GP_{(x,y)},(h,r,?)})\\
&=\frac{\sum_{i=1}^k\sum_{t_j \in a_{(h,r,?)}}drank_{i,j}}{{k}}
\end{aligned}
\end{equation*}
where $t_j$ is an entity represented by $j$ and $drank_{i,j}$ is an element of  $dRANK({GP_{(x,y)},(h,r,?)})$.
The elements related with correct entities ranked higher or equal to $k$ are summed up as the traditional precision at $k$.

Then, the {\it distributed average precision}, $dAP$, is defined for a pattern $GP_{(x,y)}$ and a training query $(h,r,?)$ as follows:	
\begin{equation*}
\begin{aligned}
&dAP({GP_{(x,y)},(h,r,?)})
\\&=\frac{\sum_{t_j \in a_{(h,r,?)}}\sum_{k=1}^ndPre_k(GP_{(x,y)},(h,r,?)))\times drank_{k,j} }{|a_{(h,r,?)}|}
\end{aligned}
\end{equation*}
where $t_j$ is an entity represented by $j$, $drank_{i,j}$ is an element of  $dRANK({GP_{(x,y)},(h,r,?)})$, and $n$ is the number of entities.
The numerator of the average precision for traditional rankings is the summation of the precision at $k$ for relevant entities.
However, a relevant entity represented by $j$ is distributed over multiple ranks in $dRANK$ so that the precision at $k$ multiplied by $drank_{k,j}$ is summed over $k$ where a relevant entity $j$ is distributed.
$dAP({GP_{(x,y)},(?,r,t)})$ for a training query with a missing head can be defined in the same way. The {\it distributed mean average precision} for a GPAR $GP_{(x,y)}\Rightarrow r_{(x,y)}$ is defined as follows:
\begin{eqnarray*}
	\begin{aligned}
		&{\it dMAP_{head}}(GP_{(x,y)}\Rightarrow r_{(x,y)})
		\\&=\sum_{(?,r,t)\in Q_{r,{\rm head}}}\frac{dAP({GP_{(x,y)},(?,r,t)})}{|Q_{r,{\rm head}}|}
		\\
		&{\it dMAP_{tail}}(GP_{(x,y)}\Rightarrow r_{(x,y)})
		\\ &=\sum_{(h,r,?)\in Q_{r,{\rm tail}}}\frac{dAP({GP_{(x,y)},(h,r,?)})}{|Q_{r,{\rm tail}}|}
	\end{aligned}
\end{eqnarray*}
We also define dMAP with for the ``{\it filtered}" \cite{DBLP:conf/nips/BordesUGWY13} rankings which are obtained from original rankings by eliminating entities whose corresponding triples (except the target triple) were included in the training dataset.
"Filtered" dMAP (fdMAP) is the mean of the dAP of ''filtered" rankings for each answer of queries.

We refer to GPARs considered as entity ranking systems with these dMAPs or fdMAPs as the {\it graph pattern entity ranking model} (GRank).
\begin{table}[tb]
	\caption{Statistics of benchmark datasets. The numbers of entities, relations, training triples, validation triples, and test triples are shown.}
	\label{Table 2}
	\begin{center}
		\scalebox{0.65}{
			\begin{tabular}{c|ccccc}
				Dataset&\# Entities&\# Relations&\# Training &\# Validation&\# Test\\\hline 
				WN18&40,943&18&141,442&5,000&5,000\\
				WN18RR&40,943&11&86,835&3,034&3,134\\
				FB15k&14,951&1,345&483,142&50,000&59,071\\
				FB15k-237&14,541&237&272,115&17,535&20,466\\

			\end{tabular}
		}
	\end{center}
\end{table}
\begin{table*}[tb]
	\caption{Mean Reciprocal Rank (MRR) and HITS@\textit{n} scores obtained for the link prediction tasks on the WN18, FB15k, WN18RR, and FB15k-237 datasets. 
		The highest result for each column is shown in bold.
		The results of TransE and TorusE were reported by \citeauthor{DBLP:journals/corr/abs-1711-05435} \shortcite{DBLP:journals/corr/abs-1711-05435},
		the results of RESCAL were reported by \protect\citeauthor{DBLP:conf/aaai/NickelRP16} \shortcite{DBLP:conf/aaai/NickelRP16}, 
		the results of DistMult and ComplEx were reported by \protect\citeauthor{DBLP:conf/icml/TrouillonWRGB16} \shortcite{DBLP:conf/icml/TrouillonWRGB16}, 
		the results of R-GCN and ConvE were reported by \protect\citeauthor{DBLP:journals/corr/DettmersMSR17} \shortcite{DBLP:journals/corr/DettmersMSR17},
		the results of PRA were reported by \citeauthor{Liu:2016:HRW:2911451.2911509} \shortcite{Liu:2016:HRW:2911451.2911509},
		and the results of Node+LinkFeat were reported by \citeauthor{observed-versus-latent-features-for-knowledge-base-and-text-inference} \shortcite{observed-versus-latent-features-for-knowledge-base-and-text-inference}. 
	}
	\label{result1} 
	\label{result2}
	\begin{center}
		\scalebox{0.65}{
			\begin{tabular}{ccccccccccccccccc} \hline
				&\multicolumn{4}{c}{WN18}&\multicolumn{4}{c}{FB15k}&\multicolumn{4}{c}{WN18RR}&\multicolumn{4}{c}{FB15k-237}\\
				\cmidrule(rl){2-5}\cmidrule(rl){6-9}\cmidrule(rl){10-13}\cmidrule(rl){14-17}
				&\multicolumn{1}{c}{MRR}&\multicolumn{3}{c}{HITS@}&\multicolumn{1}{c}{MRR}&\multicolumn{3}{c}{HITS@}&\multicolumn{1}{c}{MRR}&\multicolumn{3}{c}{HITS@}&\multicolumn{1}{c}{MRR}&\multicolumn{3}{c}{HITS@}\\
				\cmidrule(rl){3-5}\cmidrule(rl){7-9}\cmidrule(rl){11-13}\cmidrule(rl){15-17}
				Model&&{1}&{3}&{10}&&{1}&{3}&{10}&&{1}&{3}&{10}&&{1}&{3}&{10}\\\hline 
				TransE&0.397&0.040&0.745&0.923&0.414&0.247&0.534&0.688&0.182&0.027&0.295&0.444&0.257&0.174&0.284&0.420\\
				TorusE&{0.947}&{0.943}&{0.950}&0.954&0.733&0.674&0.771&0.832&--&--&--&--&--&--&--&--\\
				RESCAL&0.890&0.842&0.904&0.928&0.354&0.235&0.409&0.587&--&--&--&--&--&--&--&--\\
				DistMult&0.822&0.728&0.914&0.936&0.654&0.546&0.733&0.824&0.43&0.39&0.44&0.49&0.241&0.155&0.263&0.419\\
				ComplEx&0.941&0.936&0.945&0.947&0.692&0.599&0.759&0.840&0.44&0.41&0.46&{0.51}&0.240&0.152&0.263&0.419\\
				R-GCN&0.814&0.686&0.928&{0.955}&0.651&0.541&0.736&0.825&--&--&--&--&0.248&0.153&0.258&0.417\\
				ConvE&0.942&0.935&0.947&{0.955}&0.745&0.670&0.801&{0.873}&{0.46}&0.39&0.43&0.48&{0.316}&\textbf{0.239}&{0.350}&\textbf{0.491}\\
				PRA&0.458&0.422&--&0.481&0.336&0.303&--&0.392&--&--&--&--&--&--&--&--\\
				Node+LinkFeat&0.940&--&--&0.943&0.822&--&--&0.870&--&--&--&--&0.272&--&--&0.414\\\hline
				GPro&\textbf{0.950}&\textbf{0.946}&\textbf{0.954}&\textbf{0.959}&0.793&0.759&0.810&0.858&{0.467}&{0.430}&\textbf{0.485}&\textbf{0.543}&0.229&0.163&0.250&0.360\\
				GRank (dMAP)&\textbf{0.950}&\textbf{0.946}&0.953&0.957&{0.841}&{0.814}&{0.855}&{0.890}&{0.466}&{0.434}&{0.480}&{0.530}&0.312&0.233&0.340&0.473\\
				GRank(fdMAP)&\textbf{0.950}&\textbf{0.946}&\textbf{0.954}&0.958&\textbf{0.842}&\textbf{0.816}&\textbf{0.856}&\textbf{0.891}&\textbf{0.470}&\textbf{0.437}&{0.482}&{0.539}&\textbf{0.322}&\textbf{0.239}&\textbf{0.352}&0.489\\\hline

			\end{tabular}
		}
	\end{center}
\end{table*}

By using a graph pattern to rank entities, GRank is able to properly estimate GPARs where the number of matches is important as shown in Section \ref{problems}, unlike GPro.
For example, ${\it dMAP_{tail}}(AR_{1,(x,y)})=1$, is the maximum value, while ${\it dMAP_{tail}}(AR_{2,(x,y)})=1/2$ in Figure \ref{graph}.
Hence, GRank can answer the query $({\rm Team\ C},{\sf located\_in},?)$ by applying $AR_{1,(x,y)}$. 

\section{Experiments}
\label{experiments}
Our proposed models, GPro (Section \ref{problems}) and GRank (Section \ref{proposing}), are evaluated through link prediction tasks and compared with other state-of-the-art link prediction models.

\subsection{Datasets}
Experiments were conducted on four benchmark datasets: WN18, FB15k \cite{DBLP:conf/nips/BordesUGWY13}, WN18RR \cite{DBLP:journals/corr/DettmersMSR17}, and FB15k-237 \cite{observed-versus-latent-features-for-knowledge-base-and-text-inference} (details of these datasets are provided in Table \ref{Table 2}).
These datasets have been widely used in previous studies for evaluating model performance in link prediction tasks.

WN18 and FB15k were extracted from the real knowledge graphs WordNet \cite{Miller:1995:WLD:219717.219748} and Freebase \cite{Bollacker:2008:FCC:1376616.1376746}, respectively.
WordNet is a well-known human-curated lexical database, and
hence, WN18 is an easy benchmark of link prediction because it is well constructed and there are few missing or wrong facts.
Therefore, link prediction models should perform well on WN18.
Freebase is a huge knowledge graph of general facts and there are many missing facts.
It is known that WN18 and FB15k have redundancy in the form of reverse relations.  
For this reason, when WN18RR and FB15k-237 are extracted from WN18 and FB15k, the inverse relations of other relations are removed. 



\subsection{Evaluation Protocol}
We conducted the link prediction task following the same approach reported in \cite{DBLP:conf/nips/BordesUGWY13} to evaluate our models qualitatively and quantitatively.
For each test triple $(h_t,r_t,t_t)$ in a dataset, two queries, $(h_t,r_t,?)$ and $(?,r_t,t_t)$, were constructed in the same way as in Section \ref{reconstruction}.
Then, we obtained the rankings of entities for each query from each model as outlined in the following paragraphs.
The rankings were ''filtered" by eliminating entities whose corresponding triples (except the target test triple) were included in the training, validation, or test dataset.
The obtained rankings were scored by their mean reciprocal rank (MRR) and HITS@\textit{n},
where MRR is the mean of the inverse of the ranks of corresponding entities and HITS@\textit{n} is the proportion of test queries whose corresponding entities are ranked in the top \textit{n} of the obtained rankings.

Next, we describe how to obtain rankings from models. We restricted antecedent graph patterns of GPARs to {\it connected} and {\it closed} \cite{DBLP:conf/www/GalarragaTHS13,DBLP:journals/vldb/GalarragaTHS15} patterns whose size ${|GP_{(x,y)}|}$ was less than or equal to $L$ to restrict the search space.
A connected and closed patterns is a pattern connecting $x$ and $y$ without branches, as shown in Figure \ref{pattern}.
$L$ was chosen for each model among $\{1,2,3\}$ by MRR from the validation triples of each dataset.
It took about four days to evaluate all candidate GPARs for GRank with dMAPs in FB15k using an {\it Intel Xeon Gold 6154} (3.00 GHz, 18 cores).

We now explain how we obtained the rankings for queries with missing heads. 
For each relation $r$, we chose the top 1,000 GPARs in descending order of the standard confidence, the dMAP, or the fdMAP to predict the heads.
Let $GP_{i,(x,y)}\Rightarrow r_{(x,y)}$ be the obtained GPAR, where $i$ denotes the rank.
We defined the ordering for two entities for query $(?,r_t,t_t)$ as follows: for entities $e_1$ and $e_2$, we define $e_1 >e_2$ if there exists $i'$ for which $score(GP_{i,(x,y)},(e_1,t_t))=score(GP_{i,(x,y)},(e_2,t_t))$ for $i>i'$ and $score(GP_{i',(x,y)},(e_1,t_t))>score(GP_{i',(x,y)},(e_2,t_t))$.
We obtained the entity rankings with this ordering for each query.
Rankings for queries with missing tails were obtained in the same way.

\begin{figure*}[tb]
	\centering
	\includegraphics[scale=1,trim=25 0 0 0]{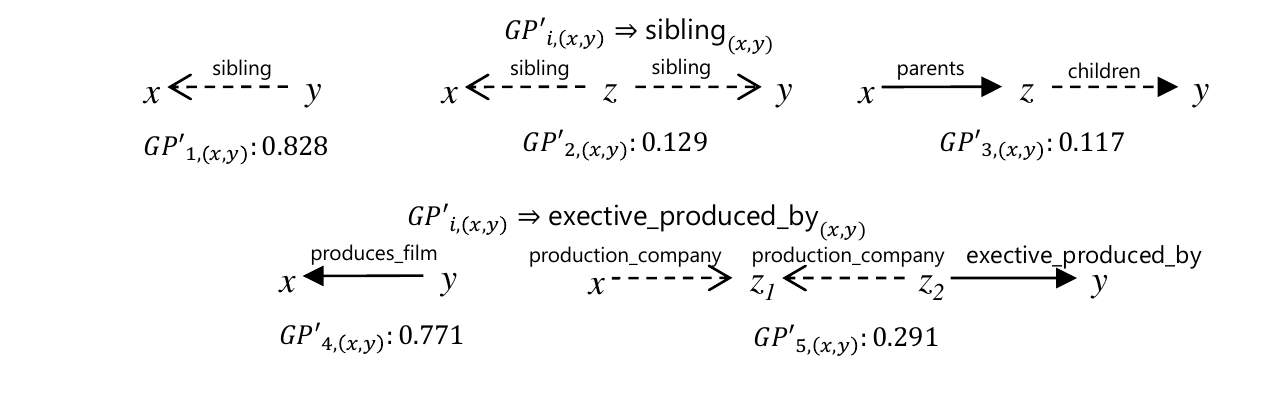}
	\caption{Examples of antecedent patterns for ${\it dMAP_{tail}}$ which were given high ranks by GRank for the FB15k dataset.
	}
	\label{patternFB}
\end{figure*}
\subsection{Results}

The results of the link prediction tasks for our proposed models, GPro, GRank with dMAP, and GRank with fdMAP, are shown in Tables \ref{result1}, where the results reported in previous studies are included for comparison.

In Table \ref{result1}, the first seven models are knowledge graph embedding models and the following two models are observed feature models.
Table \ref{result1} shows the effectiveness of the Node+LinkFeat model \cite{observed-versus-latent-features-for-knowledge-base-and-text-inference}, although this model is very simple (high MRRs imply that the model also has high HITS@1s or HITS@3s).
The Node+LinkFeat model performed well on WN18 and FB15k because these datasets often contain the reverse relations of other relations.
In other words, it shows that knowledge graph embedding models failed to capture this redundancy.
On the other hand, 	our proposed models, GPro and GRank, generally yield better results than the knowledge graph embedding models and results which are better than or comparable to Node+LinkFeat, which means that our models can also handle such redundancy.
In particular, GRank with dMAP and fdMAP yielded the best results on FB15k.
This indicates that taking the multiplicity of matchings and deeper information into account is important for knowledge graphs such as FreeBase that contain miscellaneous relations and are not well curated like WordNet.
As a result, GRank performed well.

Table \ref{result2} also shows GPro and GRank yield better results for the WN18RR dataset than the other models.
For FB15k-237, the performance of Node+LinkFeat is comparable with most of the other more sophisticated knowledge graph models and GPro does not yield good results because FB15k-237 has less redundancy.
GRank also performs better than the most other models for the FB15k-237 dataset for the same reason as the FB15k dataset.
However, our models do not utilize the information related to the co-occurrence of entities and relations in triples (node features \cite{observed-versus-latent-features-for-knowledge-base-and-text-inference}), while ConvE, Node+LinkFeat, and other models do.
We also limited the size and the shapes of graph patterns because of the calculation time; we will address these and improve our models further in our future work.

\paragraph{Quality of Obtained Paths}

The examples of antecedent patterns ranked high by GRank with ${\it dMAP_{tail}}$ for FB15k are shown in Figure \ref{patternFB}.
The patterns shown for predicting the {\sf sibling} relation are all correct as the antecedents of GPARs; however, the MAP of $GP'_{2,(x,y)}$ and $GP'_{3,(x,y)}$ are low.
The reason for this is that $GP'_{2,(x,y)}$ works when an individual has more than two siblings.
The MAP of $GP'_{3,(x,y)}$ is low because individual's parents are often missing in FB15k.
However, they are still ranked higher than other patterns.

The {\sf produces\_film} relation is the inverse relation of the {\sf exective\_produced\_by} relation in FB15k.
Such patterns are very helpful when performing link prediction tasks, and GRank is able to find them.
However, the MAP is not as high because of missing facts.
GRank is able to use majority rules such as $GP'_{5,(x,y)}\Rightarrow {\sf film\_ produced\_ by(x,y)}$ instead in such cases.
This rule can be interpreted as stating that a particular film was likely to have been produced by a person who produced many films in the same production company.

Output triples of GRank (and GPAR-based models) are described by antecedent patterns unlike knowledge graph embedding models as shown here.

\section{Conclusions}
\label{conclusions}
In this paper, we first defined GPARs for a knowledge graph and the standard confidence measures of GPARs for link prediction.
Then, we pointed out the problems with the standard confidence measures and we introduced a new perspective using GPARs to rank entities to overcome these problems.
We also proposed distributed rankings for situations where multiple entities have the same scores and defined metrics for them.
This idea led us to propose the GRank model. 
GRank is easy to interpret because outputs are described by GPARs, unlike knowledge graph embedding models, and so efficient that it outperformed the state-of-the-art knowledge graph embedding models in link prediction tasks.

In future work, we will extend GRank to use more complex patterns.
We considered only antecedent graph patterns whose sizes were less than or equal to 3.
If we allow antecedent graph patterns to have larger sizes, then we may find more useful GPARs.
We also restricted graph patterns to contain only variables and not constants.
Hence, we did not use all of the available information contained in the knowledge graph.
We believe that using such complex graph patterns will improve GRank further.

\section*{Acknowledgements}
This work was partially supported by the New Energy and Industrial Technology Development Organization (NEDO).

We would like to thank Patrik Schneider for helpful writing advice.

\bibliographystyle{acl_natbib}
\bibliography{bib2}

\begin{thebibliography}{26}
\expandafter\ifx\csname natexlab\endcsname\relax\def\natexlab#1{#1}\fi

\bibitem[{Auer et~al.(2007)Auer, Bizer, Kobilarov, Lehmann, Cyganiak, and
  Ives}]{DBLP:conf/semweb/AuerBKLCI07}
S{\"{o}}ren Auer, Christian Bizer, Georgi Kobilarov, Jens Lehmann, Richard
  Cyganiak, and Zachary~G. Ives. 2007.
\newblock {DB}pedia: {A} nucleus for a web of open data.
\newblock In \emph{The Semantic Web, 6th International Semantic Web Conference,
  2nd Asian Semantic Web Conference}, pages 722--735.

\bibitem[{Bollacker et~al.(2008)Bollacker, Evans, Paritosh, Sturge, and
  Taylor}]{Bollacker:2008:FCC:1376616.1376746}
Kurt Bollacker, Colin Evans, Praveen Paritosh, Tim Sturge, and Jamie Taylor.
  2008.
\newblock Freebase: A collaboratively created graph database for structuring
  human knowledge.
\newblock In \emph{Proceedings of the 2008 ACM SIGMOD International Conference
  on Management of Data}, pages 1247--1250.

\bibitem[{Bordes et~al.(2014)Bordes, Chopra, and Weston}]{D14-1067}
Antoine Bordes, Sumit Chopra, and Jason Weston. 2014.
\newblock Question answering with subgraph embeddings.
\newblock In \emph{Proceedings of the 2014 Conference on Empirical Methods in
  Natural Language Processing (EMNLP)}, pages 615--620.

\bibitem[{Bordes et~al.(2013)Bordes, Usunier, Garc{\'{\i}}a{-}Dur{\'{a}}n,
  Weston, and Yakhnenko}]{DBLP:conf/nips/BordesUGWY13}
Antoine Bordes, Nicolas Usunier, Alberto Garc{\'{\i}}a{-}Dur{\'{a}}n, Jason
  Weston, and Oksana Yakhnenko. 2013.
\newblock Translating embeddings for modeling multi-relational data.
\newblock In \emph{Advances in Neural Information Processing Systems}, pages
  2787--2795.

\bibitem[{Daiber et~al.(2013)Daiber, Jakob, Hokamp, and
  Mendes}]{Daiber:2013:IEA:2506182.2506198}
Joachim Daiber, Max Jakob, Chris Hokamp, and Pablo~N. Mendes. 2013.
\newblock Improving efficiency and accuracy in multilingual entity extraction.
\newblock In \emph{Proceedings of the 9th International Conference on Semantic
  Systems}, pages 121--124.

\bibitem[{Defferrard et~al.(2016)Defferrard, Bresson, and
  Vandergheynst}]{NIPS2016_6081}
Micha\"{e}l Defferrard, Xavier Bresson, and Pierre Vandergheynst. 2016.
\newblock Convolutional neural networks on graphs with fast localized spectral
  filtering.
\newblock In \emph{Advances in Neural Information Processing Systems 29}, pages
  3844--3852.

\bibitem[{Dettmers et~al.(2018)Dettmers, Minervini, Stenetorp, and
  Riedel}]{DBLP:journals/corr/DettmersMSR17}
Tim Dettmers, Pasquale Minervini, Pontus Stenetorp, and Sebastian Riedel. 2018.
\newblock Convolutional 2d knowledge graph embeddings.
\newblock In \emph{Proceedings of the Thirtieth {AAAI} Conference on Artificial
  Intelligence}.

\bibitem[{Duvenaud et~al.(2015)Duvenaud, Maclaurin, Iparraguirre, Bombarell,
  Hirzel, Aspuru-Guzik, and Adams}]{NIPS2015_5954}
David~K Duvenaud, Dougal Maclaurin, Jorge Iparraguirre, Rafael Bombarell,
  Timothy Hirzel, Alan Aspuru-Guzik, and Ryan~P Adams. 2015.
\newblock Convolutional networks on graphs for learning molecular fingerprints.
\newblock In \emph{Advances in Neural Information Processing Systems 28}, pages
  2224--2232.

\bibitem[{Ebisu and Ichise(2018)}]{DBLP:journals/corr/abs-1711-05435}
Takuma Ebisu and Ryutaro Ichise. 2018.
\newblock Toruse: Knowledge graph embedding on a lie group.
\newblock In \emph{Proceedings of the Thirtieth {AAAI} Conference on Artificial
  Intelligence}.

\bibitem[{Fan et~al.(2015)Fan, Wang, Wu, and Xu}]{DBLP:journals/pvldb/FanWWX15}
Wenfei Fan, Xin Wang, Yinghui Wu, and Jingbo Xu. 2015.
\newblock Association rules with graph patterns.
\newblock \emph{{VLDB} J.}, 8(12):1502--1513.

\bibitem[{Gal{\'{a}}rraga et~al.(2015)Gal{\'{a}}rraga, Teflioudi, Hose, and
  Suchanek}]{DBLP:journals/vldb/GalarragaTHS15}
Luis Gal{\'{a}}rraga, Christina Teflioudi, Katja Hose, and Fabian~M. Suchanek.
  2015.
\newblock Fast rule mining in ontological knowledge bases with {AMIE+}.
\newblock \emph{{VLDB} J.}, 24(6):707--730.

\bibitem[{Gal{\'{a}}rraga et~al.(2013)Gal{\'{a}}rraga, Teflioudi, Hose, and
  Suchanek}]{DBLP:conf/www/GalarragaTHS13}
Luis~Antonio Gal{\'{a}}rraga, Christina Teflioudi, Katja Hose, and Fabian~M.
  Suchanek. 2013.
\newblock {AMIE:} association rule mining under incomplete evidence in
  ontological knowledge bases.
\newblock In \emph{22nd International World Wide Web Conference}, pages
  413--422.

\bibitem[{Hakimov et~al.(2012)Hakimov, Oto, and
  Dogdu}]{Hakimov:2012:NER:2237867.2237871}
Sherzod Hakimov, Salih~Atilay Oto, and Erdogan Dogdu. 2012.
\newblock Named entity recognition and disambiguation using linked data and
  graph-based centrality scoring.
\newblock In \emph{Proceedings of the 4th International Workshop on Semantic
  Web Information Management}, SWIM '12, pages 1--7.

\bibitem[{Kipf and Welling(2017)}]{DBLP:journals/corr/KipfW16}
Thomas~N. Kipf and Max Welling. 2017.
\newblock \href {http://arxiv.org/abs/1609.02907} {Semi-supervised
  classification with graph convolutional networks}.
\newblock In \emph{Proceedings of the Fifth International Conference on
  Learning Representations}.

\bibitem[{Lao and Cohen(2010)}]{DBLP:journals/ml/LaoC10}
Ni~Lao and William~W. Cohen. 2010.
\newblock \href {https://doi.org/10.1007/s10994-010-5205-8} {Relational
  retrieval using a combination of path-constrained random walks}.
\newblock \emph{Machine Learning}, 81(1):53--67.

\bibitem[{Lao et~al.(2011)Lao, Mitchell, and Cohen}]{DBLP:conf/emnlp/LaoMC11}
Ni~Lao, Tom~M. Mitchell, and William~W. Cohen. 2011.
\newblock \href {http://www.aclweb.org/anthology/D11-1049} {Random walk
  inference and learning in {A} large scale knowledge base}.
\newblock In \emph{Proceedings of the 2011 Conference on Empirical Methods in
  Natural Language Processing}, pages 529--539.

\bibitem[{Liu et~al.(2016)Liu, Jiang, Han, Liu, and
  Qin}]{Liu:2016:HRW:2911451.2911509}
Qiao Liu, Liuyi Jiang, Minghao Han, Yao Liu, and Zhiguang Qin. 2016.
\newblock \href {https://doi.org/10.1145/2911451.2911509} {Hierarchical random
  walk inference in knowledge graphs}.
\newblock In \emph{Proceedings of the 39th International ACM SIGIR Conference
  on Research and Development in Information Retrieval}, pages 445--454.

\bibitem[{Miller(1995)}]{Miller:1995:WLD:219717.219748}
George~A. Miller. 1995.
\newblock Wordnet: A lexical database for {E}nglish.
\newblock \emph{Commun. ACM}, 38(11):39--41.

\bibitem[{Nickel et~al.(2016)Nickel, Rosasco, and
  Poggio}]{DBLP:conf/aaai/NickelRP16}
Maximilian Nickel, Lorenzo Rosasco, and Tomaso~A. Poggio. 2016.
\newblock Holographic embeddings of knowledge graphs.
\newblock In \emph{Proceedings of the Thirtieth {AAAI} Conference on Artificial
  Intelligence}, pages 1955--1961.

\bibitem[{Nickel et~al.(2011)Nickel, Tresp, and
  Kriegel}]{DBLP:conf/icml/NickelTK11}
Maximilian Nickel, Volker Tresp, and Hans{-}Peter Kriegel. 2011.
\newblock A three-way model for collective learning on multi-relational data.
\newblock In \emph{Proceedings of the 28th International Conference on Machine
  Learning}, pages 809--816.

\bibitem[{Schlichtkrull et~al.(2017)Schlichtkrull, Kipf, Bloem, van~den Berg,
  Titov, and Welling}]{DBLP:journals/corr/SchlichtkrullKB17}
Michael~Sejr Schlichtkrull, Thomas~N. Kipf, Peter Bloem, Rianne van~den Berg,
  Ivan Titov, and Max Welling. 2017.
\newblock \href {http://arxiv.org/abs/1703.06103} {Modeling relational data
  with graph convolutional networks}.
\newblock \emph{CoRR}, abs/1703.06103.

\bibitem[{Socher et~al.(2013)Socher, Chen, Manning, and
  Ng}]{DBLP:conf/nips/SocherCMN13}
Richard Socher, Danqi Chen, Christopher~D. Manning, and Andrew~Y. Ng. 2013.
\newblock Reasoning with neural tensor networks for knowledge base completion.
\newblock In \emph{Advances in Neural Information Processing Systems}, pages
  926--934.

\bibitem[{Suchanek et~al.(2007)Suchanek, Kasneci, and
  Weikum}]{DBLP:conf/www/SuchanekKW07}
Fabian~M. Suchanek, Gjergji Kasneci, and Gerhard Weikum. 2007.
\newblock Yago: a core of semantic knowledge.
\newblock In \emph{Proceedings of the 16th International Conference on World
  Wide Web}, pages 697--706.

\bibitem[{Toutanova and
  Chen(2015)}]{observed-versus-latent-features-for-knowledge-base-and-text-inference}
Kristina Toutanova and Danqi Chen. 2015.
\newblock Observed versus latent features for knowledge base and text
  inference.
\newblock In \emph{Proceedings of the 3rd Workshop on Continuous Vector Space
  Models and Their Compositionality}.

\bibitem[{Trouillon et~al.(2016)Trouillon, Welbl, Riedel, Gaussier, and
  Bouchard}]{DBLP:conf/icml/TrouillonWRGB16}
Th{\'{e}}o Trouillon, Johannes Welbl, Sebastian Riedel, {\'{E}}ric Gaussier,
  and Guillaume Bouchard. 2016.
\newblock Complex embeddings for simple link prediction.
\newblock In \emph{Proceedings of the 33rd International Conference on Machine
  Learning}, pages 2071--2080.

\bibitem[{Yang et~al.(2015)Yang, Yih, He, Gao, and
  Deng}]{DBLP:journals/corr/YangYHGD14a}
Bishan Yang, Wen{-}tau Yih, Xiaodong He, Jianfeng Gao, and Li~Deng. 2015.
\newblock Embedding entities and relations for learning and inference in
  knowledge bases.
\newblock In \emph{Proceedings of the Third International Conference on
  Learning Representations}.

\end{thebibliography}

\end{document}